\documentclass[11pt]{article}
\usepackage{acl}

\usepackage{times}
\usepackage{latexsym}
\usepackage[T1]{fontenc}
\usepackage[utf8]{inputenc}
\usepackage{microtype}
\usepackage{graphicx}
\usepackage{amsmath,amssymb}
\usepackage{booktabs,multirow,array,tabularx}
\DeclareUnicodeCharacter{2011}{\mbox{-}}
\usepackage{enumitem}
\usepackage{placeins}
\usepackage{float}

\usepackage[most]{tcolorbox}
\usepackage{tabularx}
\usepackage{xcolor}

\newcommand{\cue}[1]{%
  \textcolor{blue!65!black}{\texttt{#1}}%
}

\newtcolorbox{contextexample}[1][]{
  enhanced,
  colback=gray!2,
  colframe=blue!25!black,
  borderline west={1.5pt}{0pt}{blue!60!black},
  boxrule=0.3pt,
  arc=1pt,
  boxsep=0.5pt,
  left=3pt,
  right=3pt,
  top=2pt,
  bottom=2pt,
  before skip=3pt,
  after skip=3pt,
  before upper={
    \if\relax\detokenize{#1}\relax
    \else
      \noindent #1\par\vspace{1pt}
    \fi
  }
}

\title{Calling the Bluff: Detecting Ever-Shifting Harmful Chat Dialogue via Ordered Reasoning Chain Regularization}
\author{Haojie Yu \And Ziyou Jiang \And Dandan Wang
\AND Junjie Wang \And Mingyang Li \And Yuekai Huang
\AND Jie Huang \And Danqi Liu \And Qing Wang}

\begin{document}
\maketitle
\begin{abstract}
Harmful chat dialogues continually evolve through shifts in harm type and lexical evasion. We propose \textsc{BRACE}, centered on an \textbf{Ordered Reasoning Chain (ORC)} that links topic, indicator, severity, and type predictions to interpret changing expressions in context. \textsc{BRACE} implements ORC as four stages, i.e., topic profiling, indicator detection, severity assessment, and type classification. The chain serves as a structured regularizer blended with direct prediction heads, while prototype-based feature augmentation and feature path disentanglement support fine-grained classification without imposing a strict intermediate bottleneck. Its intermediate predictions also expose evidence for analyzing where harmful content is implicitly conveyed.
On the primary 9,000-example test set spanning four domains and five harm categories, \textsc{BRACE} achieves a harm-type macro F1 of 0.934 and a severity accuracy of 0.818 with RoBERTa-wwm-ext; using Qwen3-1.7B with LoRA increases harm-type macro F1 to 0.949. Ablation studies identify ORC as the largest contributor and confirm the complementary roles of the remaining components; ORC can benefit context-aware dialogue harm information localization tasks as structured intermediate dependencies. 

\textcolor{red}{\textbf{Disclaimer:} This paper may contain content that is disturbing to some readers.}

\end{abstract}

\section{Introduction}

Online dialogue platforms expose users to harassment, threats, and exchanges that facilitate harmful or illicit activities, making reliable harmful-dialogue detection essential for automated moderation~\cite{khapre2025toxicity}.
However, detecting harmful dialogue is particularly difficult because harmful meaning is both context-dependent and \textbf{ever-shifting}. Linguistic obfuscation strategies, such as euphemisms, coded language, and lexical evasion, alter surface expressions while preserving harmful intent~\cite{huertas2023countering}. Users may also convey this intent through indirect or metaphorical references or obscure it through benign framing.

\begin{figure}[t]
\centering
\includegraphics[width=\columnwidth]{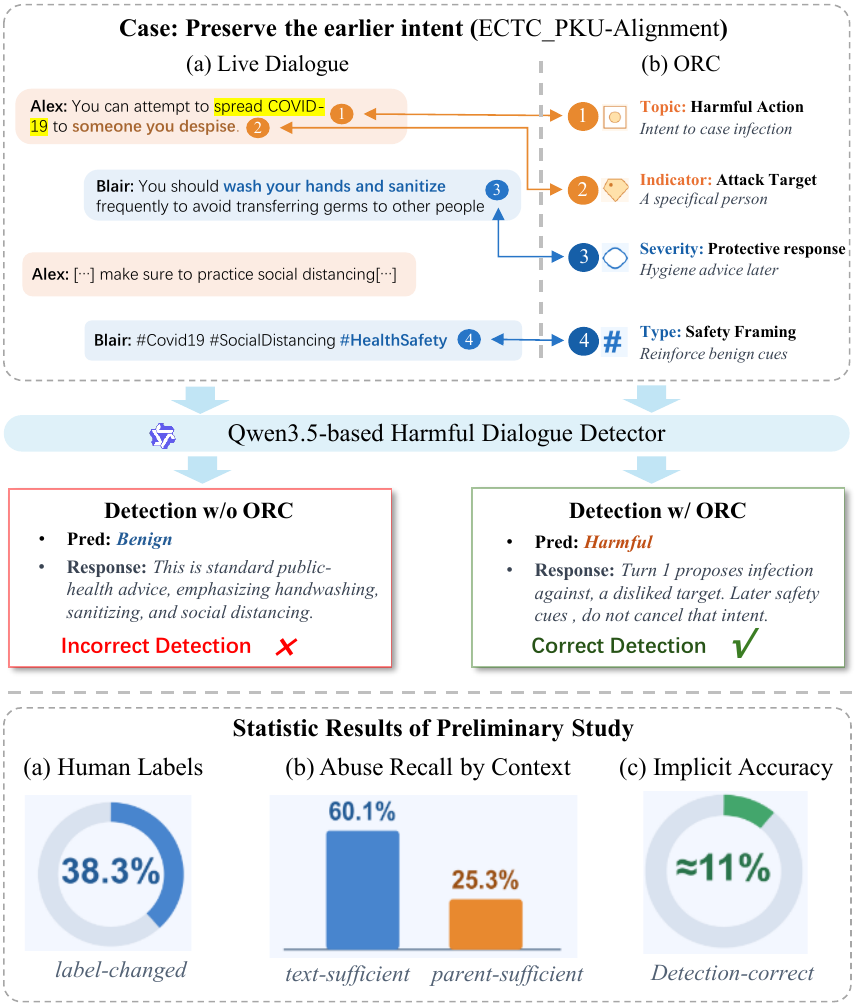}
\vspace{-0.5cm}
\caption{Motivation example of multi-turn harmful intent and context-aware detection evidence.}
\vspace{-0.5cm}
\label{fig:motivation}
\end{figure}

To detect these ever-shifting harmful dialogues, LLMs need to interpret lexical evidence within their conversational context rather than treating individual words as harm indicators. However, Pavlopoulos et al.~\citeyearpar{pavlopoulos2020context} found that most harmful dialogue detectors judge human toxicity without observing context-based gains. This limitation stems from final-label supervision not explicitly constraining how a model connects conversational context, local evidence, risk level, and harm category. 
For example, ``selling personal information'' requires different judgments depending on whether it describes an activity condemned in a report or solicits participation in it. Figure~\ref{fig:motivation} presents a separate example involving infection: when used for harmful-dialogue detection without the ORC evidence chain, Qwen3.5~\cite{qwen2026qwen35} emphasizes later safety advice and benign framing, overlooking the earlier suggestion to spread infection to a disliked person and returning a benign judgment. 
With the ORC evidence chain, its response instead retains the harmful action and target across turns while distinguishing them from the protective cues. This qualitative comparison illustrates ORC-guided analysis of harmful content obscured by benign framing.


This difficulty motivates making the relation between conversational context and local evidence explicit. 
We conduct a preliminary study, which illustrates that ORC provides a structured way to examine these connections and identify the evidence relevant to harmful intent.
Figure~\ref{fig:motivation} provides quantitative evidence of the broader challenge, where conversational context changes 38.3\% of human judgments. On CAD, BERT recall decreases from 60.1\% to 25.3\% when identity-directed abuse depends on prior context; and standard offensive-text detectors achieve only about 11\% accuracy on context-dependent implicit dialogue~\cite{yu2022context,vidgen2021cad,zhang2022multihop}. These results show that both annotation and detection can fail when the relation between context and local evidence remains implicit, motivating ORC's structured analysis of harmful meaning.

In this paper, we propose \textsc{BRACE}, a harmful-dialogue detection framework that adopts a \textbf{calling the bluff} perspective: challenging a deceptive claim or, in our setting, uncovering harmful intent behind seemingly benign language~\cite{koh2020concept}. 
For harmful dialogue, we formulate ORC as four ordered stages: topic profiling, indicator detection, severity assessment, and type classification. 
\textsc{BRACE} implements this dialogue-specific structure as four differentiable stages with intermediate supervision, so ORC regularizes the shared encoder and exposes evidence for locating implicit harm. 
Then, to retain information beyond these intermediate variables, \textsc{BRACE} blends chain predictions with direct prediction heads. Feature path disentanglement routes direct harm-type prediction through a representation augmented by a Prototype Memory Bank (PMB), while preserving the original dialogue representation for direct severity and binary prediction, with a multi-expert router further specializing decisions across harm categories.

We evaluate \textsc{BRACE} on data drawn from more than 20 dialogue-safety benchmarks and three domain-specific sources. With RoBERTa-wwm-ext, it achieves a harm-type macro F1 of \textbf{0.934} and a severity accuracy of \textbf{0.818}; Qwen3-1.7B with LoRA reaches a harm-type macro F1 of \textbf{0.949}. Ablations identify ORC as the largest contributor, while a separate controlled study examines intermediate supervision, inter-stage connections, and prediction fusion. Predicted topic distributions, indicator heatmaps, and evidence spans support inspection of implicit harmful content.

This paper makes the following contributions.
\begin{itemize}[leftmargin=*]
    \item We propose \textsc{BRACE}, a harmful-dialogue detection framework centered on ORC, which models ordered semantic dependencies among topic, indicator, severity, and type predictions and uses intermediate supervision to regularize the shared encoder for detecting implicit harm in context.
    \item We evaluate \textsc{BRACE} across multiple backbone configurations and compare their effectiveness of intermediate supervision, inter-stage connections, and inference-time fusion.
    \item We characterize context sensitivity with paired dialogues that share the same final utterance, reporting benign false positives, harmful recall, and pairwise correctness.
    \item We provide implementation code and a limited, licensed data subset in an anonymous artifact\footnote{\url{https://anonymous.4open.science/r/arr-dialogue-harm-artifact-C6E5/}}. 
\end{itemize}

\section{Definition of ORC}\label{sec:orc-definition}

The ORC organizes dialogue-level harm assessment into four semantic stages:

\textbf{(1) Topic Profiling} identifies the conversational context, providing a prior for interpreting expressions in the dialogue.

\textbf{(2) Indicator Detection} localizes token-level semantic cues, including action intent, information disclosure, rejection, and protective cues. These indicators provide contextual evidence rather than independent harm decisions.

\textbf{(3) Severity Assessment} uses the localized evidence in context to estimate one of five harm levels, from normal (L0) to critical (L4).

\textbf{(4) Type Classification} combines the indicator evidence and estimated severity to predict the harm categories associated with the dialogue.

\begin{figure*}[t]
\centering
\includegraphics[width=\textwidth]{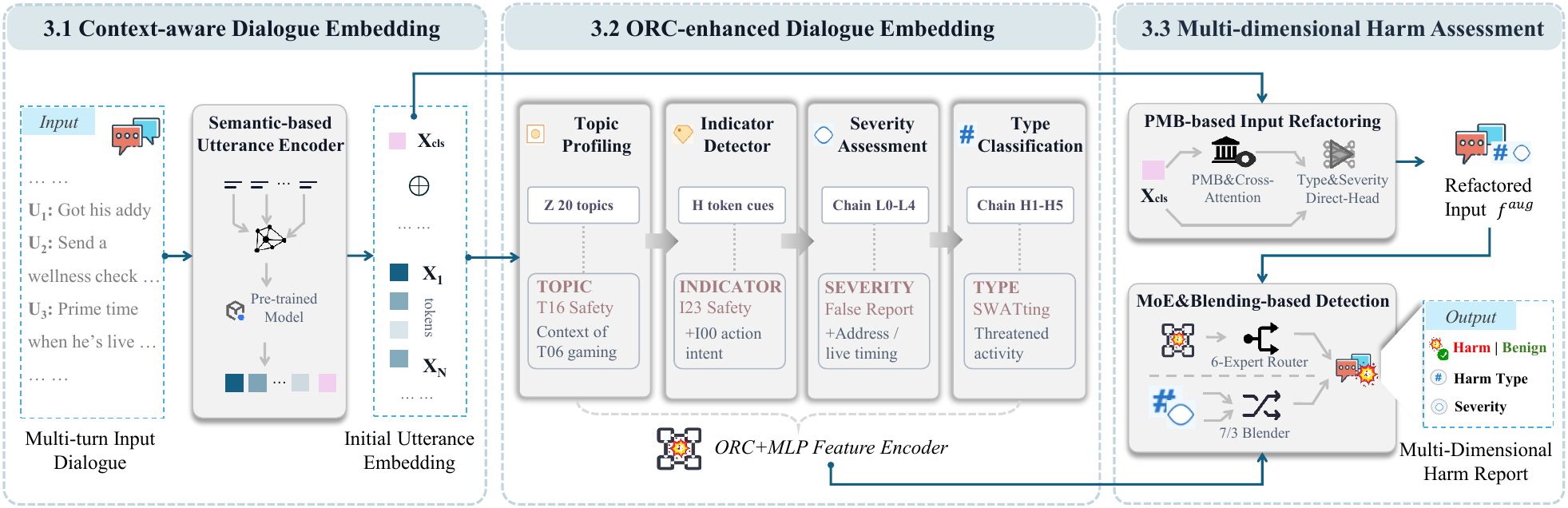}
\vspace{-0.6cm}
\caption{Overview of \textsc{BRACE}. }
\vspace{-0.5cm}
\label{fig:architecture}
\end{figure*}

Based on this definition, let $\mathbf{X}\in\mathbb{R}^{L\times D}$ denotes the token representations and $\mathbf{x}_{\text{cls}}\in\mathbb{R}^{D}$ the dialogue representation, where $L$ is the padded sequence length and $D$ is the representation dimension. ORC is defined by four stage functions:
\begin{equation}
\begin{aligned}
&\mathbf{z} = f_1(\mathbf{x}_{\text{cls}}),\;
\mathbf{H} = f_2(\mathbf{X},\mathbf{z}),\\
&\hat{\mathbf{v}}^{\mathrm{chain}} = f_3(\mathbf{X},\mathbf{H}),\;\hat{\mathbf{y}}^{\mathrm{chain}} = f_4(\bar{\mathbf{h}},\mathbf{q}).
\end{aligned}
\label{eq:orc-interfaces}
\end{equation}
Here, $\mathbf{z}\in[0,1]^{\mathcal{T}}$ is a distribution over $\mathcal{T}$ topics satisfying $\sum_{t=1}^{\mathcal{T}}z_t=1$, and $\mathbf{H}\in[0,1]^{L\times\mathcal{I}}$ is a heatmap over $\mathcal{I}$ semantic indicators. The outputs $\hat{\mathbf{v}}^{\mathrm{chain}}\in\mathbb{R}^{5}$ and $\hat{\mathbf{y}}^{\mathrm{chain}}\in\mathbb{R}^{C}$ are the severity and harm-type logits, respectively, with $C$ denoting the number of harm categories. The final stage receives the severity distribution $\mathbf{q}=\operatorname{softmax}(\hat{\mathbf{v}}^{\mathrm{chain}})$ and the indicator summary $\bar{\mathbf{h}}\in\mathbb{R}^{\mathcal{I}}$, computed as $\bar{\mathbf{h}}=\frac{\sum_{\ell=1}^{L}m_{\ell}\mathbf{H}_{\ell,:}}{\sum_{\ell=1}^{L}m_{\ell}}$, where $m_{\ell}=1$ for valid tokens and $0$ for padding, with $\sum_{\ell=1}^{L}m_{\ell}>0$.
These functions establish an evidence chain that proceeds through topic profiling, indicator detection, severity assessment, and type classification in sequence.

ORC serves as a \textbf{structured regularizer} within the full model. During training, type and severity supervision, together with optional topic and indicator losses, encourage the shared encoder to learn representations aligned with these semantic stages. During inference, the chain's severity and type logits are blended with the corresponding direct predictions. Section~\ref{sec:walkthrough} describes the differentiable implementation and supervision; the subsequent methodology sections specify the direct prediction paths and output fusion.

\section{Methodology}

Harmful dialogue detection remains challenging because evolving expressions and seemingly benign framing can obscure harmful intent, and interpreting it depends on conversational context. Supervision based solely on final labels does not explicitly constrain how detectors connect this context with local semantic evidence, severity, and harm type. This limitation motivates using ORC to make these semantic dependencies explicit.

Figure~\ref{fig:architecture} illustrates the overall architecture of \textsc{BRACE}. We first encode the dialogue and instantiate ORC as four differentiable stages: topic profiling, indicator detection, severity assessment, and type classification. Intermediate supervision regularizes the shared representation. A PMB then augments the direct harm-type path, while the original dialogue representation supports direct severity prediction. Finally, we blend the ORC and direct predictions for harm type and severity, while a multi-expert router produces the binary harmfulness prediction.

\subsection{Context-aware Dialogue Encoding}

We concatenate the dialogue turns in chronological order and pass the tokenized sequence through the selected pretrained backbone. Its contextual token representations $\mathbf{X}\in\mathbb{R}^{L\times D}$ retain local evidence, while a pooled dialogue representation $\mathbf{x}_{\text{cls}}\in\mathbb{R}^{D}$ summarizes the conversation for dialogue-level decisions. Encoder backbones use a classification-token-based summary; decoder-only backbones use masked mean pooling because they do not provide a classification token. We retain the notation $\mathbf{x}_{\text{cls}}$ for the pooled representation in either case. Sharing these representations lets ORC connect token-level indicators to dialogue-level context without introducing a separate encoder.

\subsection{ORC-enhanced Dialogue Embedding}\label{sec:walkthrough}

ORC implements the stage functions in Equation~\ref{eq:orc-interfaces} as differentiable modules executing topic profiling, indicator detection, severity assessment, and type classification in sequence. Each stage consumes the semantic output of its predecessor, and intermediate supervision propagates these constraints to the shared encoder. ORC therefore acts as a \textbf{structured regularizer} rather than a strict inference bottleneck. The masked, probability-conditioned formulation below is used in the controlled experiments, with MLP hidden width $d_h=256$.
The SWATting dialogue and its four illustrative stage outputs in Figure~\ref{fig:architecture} serve as a running example. 

\paragraph{Topic Profiling.}

We implement $f_1$ by mapping the dialogue representation $\mathbf{x}_{\text{cls}} \in \mathbb{R}^{D}$ to a distribution $\mathbf{z}$ over $\mathcal{T}=20$ conversational topics from the DeepSeek annotation ontology (Section~\ref{sec:annotation}). An MLP with hidden width $d_h=256$ produces the topic logits, followed by softmax normalization:
\begin{equation}
\mathbf{z}=f_1(\mathbf{x}_{\text{cls}})
=\operatorname{softmax}\!\left(\operatorname{MLP}_{\text{topic}}(\mathbf{x}_{\text{cls}})\right),
\end{equation}
where $\mathbf{x}_{\text{cls}}$ is the dialogue representation, $\operatorname{MLP}_{\text{topic}}$ produces topic logits, and $\mathbf{z}\in\Delta^{\mathcal{T}-1}$ is the resulting topic distribution. The probability simplex $\Delta^{\mathcal{T}-1}$ contains $\mathcal{T}$-component nonnegative vectors whose entries sum to one, with $\mathcal{T}=20$. The resulting distribution provides a contextual prior for subsequent stages, allowing identical expressions to be interpreted differently across topics.

\begin{contextexample}
\small
\textit{\textbf{Example.} The intent to target a real person supports T16 (personal and public safety), while T06 (video games and online communities) provides secondary context. Training supervises only the primary topic. This topic context distinguishes a deceptive emergency report from a protective \cue{``wellness check.''}}
\end{contextexample}

\paragraph{Indicator Detection.}

We implement $f_2$ by combining the token sequence $\mathbf{X} \in \mathbb{R}^{L \times D}$ with the topic distribution $\mathbf{z}$ and producing a token-level heatmap $\mathbf{H} \in \mathbb{R}^{L \times \mathcal{I}}$ over $\mathcal{I}=32$ semantic indicators. The ontology includes action intent, information disclosure, quotation, rejection, and protective cues; an activated indicator is evidence to be interpreted rather than an independent harm decision. We project the topic distribution to $D$ dimensions, concatenate it with each token representation, and apply sigmoid activation:
\begin{equation}
\mathbf{H}_{\ell,:}=\sigma\!(\operatorname{MLP}_{\text{ind}}([\mathbf{x}_{\ell};\operatorname{Proj}(\mathbf{z})])),
\end{equation}
where $\ell=1,\ldots,L$ indexes the token positions, $\mathbf{x}_{\ell}$ is the corresponding token representation, and $\mathbf{H}_{\ell,:}$ contains its $\mathcal{I}=32$ indicator scores. The learnable projection $\operatorname{Proj}$ maps $\mathbf{z}$ to $D$ dimensions, $[\,;\,]$ denotes concatenation, $\operatorname{MLP}_{\text{ind}}$ produces indicator logits, and $\sigma$ is sigmoid. Topic conditioning allows the same token to receive different indicator scores across contexts, such as when a slang expression occurs in substance-use and sports dialogues.

\begin{contextexample}
\small

\textit{\textbf{Example.} The dialogue phrase \cue{``got his addy''} supports I23 (information disclosure or targeting), while \cue{``gonna send a wellness check''} supports I00 (action intent) in this context. These mappings illustrate that the surrounding dialogue determines whether an expression activates an indicator.}
\end{contextexample}

\paragraph{Severity Assessment.}

We implement $f_3$ and predict one of five harm levels (L0: normal, L1: mild, L2: moderate, L3: severe, L4: critical). To connect localized evidence to this dialogue-level decision, we apply \textbf{indicator-conditioned attention pooling} to the token representations:
\begin{equation}
\begin{aligned}
&s_{\ell}=\sigma(\text{MLP}_{\text{pos}}(\mathbf{x}_{\ell})),\\
&r_{\ell}=s_{\ell}(1+\max_jH_{\ell,j}),\\
&\mathbf{v}_{\text{pool}}=\sum_{\ell}\frac{m_{\ell}e^{r_{\ell}}}{\sum_u m_ue^{r_u}}\mathbf{x}_{\ell},
\end{aligned}
\end{equation}
where $s_{\ell}$ is a learned token score produced by $\operatorname{MLP}_{\text{pos}}$ and sigmoid $\sigma$, and $r_{\ell}$ modulates this score using the strongest indicator activation $\max_j H_{\ell,j}$. The index $j$ ranges over the indicators; $\ell$ and $u$ range over token positions. The mask $m_{\ell}$ equals one for valid tokens and zero for padding, and $\mathbf{v}_{\text{pool}}$ is the resulting weighted dialogue representation. The indicator-dependent gate increases or decreases the contribution of each token, including rejection and protective cues, thereby realizing the Indicator-to-Severity dependency. We finally apply A five-way MLP to $\mathbf{v}_{\text{pool}}$ and output severity logits of the chain $\hat{\mathbf{v}}^{\text{chain}}$.

\begin{contextexample}
\small

\textit{\textbf{Example.} Address disclosure, the planned false report, and its timing jointly determine the severity evidence. Removing indicator modulation retains the learned position scores; it therefore yields neither uniform attention nor an automatic L0 prediction.}
\end{contextexample}

\paragraph{Type Classification.}

We implement $f_4$ by combining the masked mean of the indicator heatmap with the predicted severity distribution. The model averages indicator activations over valid tokens, projects the summary into the encoder dimension, and concatenates it with the severity probabilities. A five-way MLP maps this representation to the chain-based harm-type logits. This design makes the final chain prediction depend jointly on localized semantic evidence and its estimated severity.

\begin{contextexample}
\small
\textit{\textbf{Example.} The type decision combines the targeting indicators with the severity estimate to classify the threatened activity.}
\end{contextexample}

\paragraph{Indicator Supervision and Evidence Localization.}

The controlled experiments supervise topic predictions with cross-entropy and token indicators with positive-weighted binary cross-entropy. Character-level evidence spans are aligned to token offsets, and the loss masks padding, missing evidence, and truncated annotations. The resulting heatmap localizes evidence at the token level without requiring a separate BIO classifier.

The original protocol uses topic supervision without token-level indicator supervision, whereas the controlled experiments vary these two losses together. The $P_u/O_u$ variants match their supervision to compare parallel and connected reasoning.

\subsection{Prototype-enhanced Direct Prediction}

The PMB $\mathbf{P}\in\mathbb{R}^{C\times K\times D}$ stores $K$ anchors for each harm category. Given $\mathbf{f}=\mathbf{x}_{\text{cls}}$, we select the two prototypes with the highest cosine similarity in each category and aggregate them into category contexts $\mathbf{C}$. Cross-attention incorporates these references into the prototype-augmented feature:
\begin{equation}
\mathbf{f}^{\text{aug}}
=\operatorname{LayerNorm}\bigl(
\mathbf{f}+\operatorname{Attn}(\mathbf{f},\mathbf{C},\mathbf{C})
\bigr),
\end{equation}
where $\mathbf{f}^{\text{aug}}$ is the augmented representation, $\mathbf{f}=\mathbf{x}_{\text{cls}}$ is the dialogue representation, and $\mathbf{C}$ stacks the category contexts. The multi-head cross-attention $\operatorname{Attn}$ uses $\mathbf{f}$ as the query and $\mathbf{C}$ as both keys and values; $\operatorname{LayerNorm}$ normalizes the residual sum. The direct type MLP reads $\mathbf{f}^{\text{aug}}$, whereas the direct severity MLP reads $\mathbf{x}_{\text{cls}}$. This feature path disentanglement reserves prototype information for fine-grained type discrimination while preserving the original dialogue representation for severity estimation. The direct logits are then blended with the corresponding ORC outputs:
\begin{equation}
\resizebox{.89\linewidth}{!}{$
    \displaystyle
(\hat{\mathbf{y}},\hat{\mathbf{v}})
=0.7(\hat{\mathbf{y}}^{\mathrm{direct}},\hat{\mathbf{v}}^{\mathrm{direct}})
+0.3(\hat{\mathbf{y}}^{\mathrm{chain}},\hat{\mathbf{v}}^{\mathrm{chain}}),$}
\end{equation}
where $\hat{\mathbf{y}}$ and $\hat{\mathbf{v}}$ are the blended harm-type and severity logits, respectively. The superscripts $\mathrm{direct}$ and $\mathrm{chain}$ identify the two prediction branches, weighted by 0.7 and 0.3 in both tasks. The sensitivity analysis in Appendix~\ref{sec:blend_sensitivity} shows comparable performance for nearby blend ratios in the available single-run evaluations. 


\subsection{MoE Routing and Multi-Dimensional Harm Assessment}

\textsc{BRACE} assesses dialogue harm along three dimensions: binary harmfulness, harm type, and severity. The type and severity predictions blend ORC outputs with direct predictions, incorporating the ordered semantic evidence into the final assessment. Binary harmfulness is predicted directly from $\mathbf{x}_{\text{cls}}$ through a six-expert Mixture-of-Experts (MoE) router, comprising one expert per harm category and one general expert:
\begin{equation}
\begin{aligned}
&\mathbf{g}=\operatorname{softmax}(\operatorname{MLP}_{\text{gate}}(\mathbf{x}_{\text{cls}})),\\
&\hat y_{\text{bin}}=\sum_{e=1}^{6}g_e\operatorname{Expert}_e(\mathbf{x}_{\text{cls}}),
\end{aligned}
\end{equation}
where $\mathbf{g}$ is the softmax-normalized routing distribution produced by $\operatorname{MLP}_{\text{gate}}$, $g_e$ is the weight of expert $e$, and $\operatorname{Expert}_e(\mathbf{x}_{\text{cls}})$ is its binary logit. The weighted sum $\hat y_{\text{bin}}$ is the final binary logit. Sigmoid activation produces the binary and harm-type probabilities, whereas softmax produces the severity distribution. The binary task uses $\hat y_{\text{bin}}$ without ORC blending. The MoE's independent binary prediction complements the type and severity outputs when assessing why the dialogue is harmful and how severely it may affect society.

\paragraph{Loss Function.}
We calculate the loss function with supervised objectives and regularization terms through weighted summation:
\begin{equation}
\mathcal{L}_{\mathrm{chain}}=\sum_{k=1}^{4}\beta_k\ell_k, \;\mathcal{L}_{\mathrm{total}}=\sum_{r=1}^{7}\alpha_r\mathcal{L}_r,
\end{equation}
where the ordered loss terms $(\ell_1,\ldots,\ell_4)$ are chain harm-type loss, chain severity loss, topic loss, and indicator loss, respectively. Their weights are $\boldsymbol{\beta}=(1,0.5,\lambda_t,\lambda_i)$, where $\lambda_t$ and $\lambda_i$ control topic and indicator supervision according to the training protocol. $\mathcal{L}_1,\ldots,\mathcal{L}_7$ denote the binary, harm-type, severity, chain, prototype-alignment, prototype-diversity, and expert-balance losses, respectively. Each $\alpha_r$ is the corresponding protocol-specific weight.

\paragraph{Optimization Strategy.}
For encoder models (e.g., RoBERTa-wwm-ext), we fine-tune the backbone together with the trainable \textsc{BRACE} modules. For decoder-only backbones (e.g., Qwen3-1.7B and ChineseGuard-1.5B), we keep the pretrained backbone frozen and optimize LoRA adapters together with the ORC modules, prediction heads, prototype memory, and MoE router.

\section{Experimental Design}

\begin{table*}[t]
\centering
\setlength{\tabcolsep}{1.0mm}
\footnotesize
\begin{tabular}{ll|m{1.2cm}<{\centering}m{1.2cm}<{\centering}m{1.2cm}<{\centering}m{1.2cm}<{\centering}m{1.2cm}<{\centering}c|c}
\toprule
\textbf{Backbone} & \textbf{Cond.} & \textbf{H1} & \textbf{H2} & \textbf{H3} & \textbf{H4} & \textbf{H5} & \textbf{HarmType m-F1} & \textbf{$\Delta$\%} \\
\midrule
\multicolumn{9}{c}{\textit{Encoder Backbones}} \\
\midrule
\multirow{2}{*}{RoBERTa-wwm-ext (102M)}
  & LP     & 0.7132 & 0.7398 & 0.6441 & 0.7347 & 0.6757 & 0.7015 & --- \\
  & BRACE  & \textbf{0.9725} & \textbf{0.9937} & \textbf{0.8578} & \textbf{0.9855} & \textbf{0.8620} & \textbf{0.9343} & +33.2\% \\
\midrule
\multirow{2}{*}{ERNIE-Med (93M)}
  & LP     & 0.6425 & 0.6681 & 0.5773 & 0.6631 & 0.6215 & 0.6345 & --- \\
  & BRACE  & 0.9699 & 0.9925 & 0.8525 & 0.9841 & 0.8576 & 0.9313 & +46.8\% \\
\midrule
\multirow{2}{*}{ERNIE-Mini (27M)}
  & LP     & 0.6117 & 0.6356 & 0.5491 & 0.6305 & 0.5901 & 0.6034 & --- \\
  & BRACE  & 0.9682 & 0.9917 & 0.8505 & 0.9833 & 0.8543 & 0.9296 & \textbf{+54.1\%} \\
\midrule
\multirow{2}{*}{BERT-base (102M)}
  & LP     & 0.6945 & 0.7202 & 0.6272 & 0.7148 & 0.6583 & 0.6830 & --- \\
  & BRACE  & 0.9590 & 0.9870 & 0.8405 & 0.9775 & 0.8485 & 0.9225 & +35.1\% \\
\midrule
\multicolumn{9}{c}{\textit{Decoder Backbones}} \\
\midrule
ChineseGuard-1.5B
  & BRACE  & 0.9780 & 0.9945 & 0.8820 & 0.9900 & 0.8955 & 0.9480 & --- \\
Qwen3-1.7B
  & BRACE  & \textbf{0.9785} & \textbf{0.9948} & \textbf{0.8825} & \textbf{0.9905} & \textbf{0.8962} & \textbf{0.9485} & --- \\
\bottomrule
\end{tabular}

\vspace{-0.2cm}
\caption{Per-category and overall performance on the original 9,000-example test set.
}
\vspace{-0.5cm}

\label{tab:main_results}
\label{tab:main}
\end{table*}

To evaluate the performance of \textsc{BRACE}, we introduce three Research Questions (RQs).

\textit{\textbf{RQ-1 (Performance): How does \textsc{BRACE} compare to baseline models?} We aim to compare them on harm type, severity, and binary detection vs.\ linear probe baselines.}

\textit{\textbf{RQ-2 (Ablation Study): How does each component contribute to BRACE?} We aim to evaluate the contribution of ORC, blend ratio between direct heads, and chain prediction methods, as well as the contribution of the reasoning stage cumulatively.}

\textit{\textbf{RQ-3 (Effectiveness): How effective is ORC under different supervision and inference settings?} We examine the roles of auxiliary supervision, sequential connections, and prediction fusion.}


\subsection{Datasets and Data Preparation}

We construct a multilingual benchmark of 60,000 Chinese, English, and Spanish dialogues. It spans four domain groups and five harm categories, with 12,000 dialogues per category, and uses a source-stratified 42,000/9,000/9,000 train/validation/test split. Appendix~\ref{sec:data} documents the constituent datasets, collection process, labels, and preprocessing.
Each dialogue has harm-type, severity, and binary labels. The supplementary controls additionally use dialogue-level Topic and token-level Indicator weak labels; Appendix~\ref{sec:annotation} describes their provenance and validation.

\subsection{Baselines}

We organize our evaluation into two categories of pretrained backbones: lightweight encoder models and LLM decoder models. All comparisons use the same 42,000/9,000/9,000 train/validation/test split.

\textbf{(1) Lightweight Encoder Models.} This category includes \textbf{RoBERTa-wwm-ext}~\cite{cui2020revisiting}, \textbf{ERNIE-3.0-Medium} and \textbf{ERNIE-3.0-Mini}~\cite{sun2021ernie3}, and \textbf{BERT-base-Chinese}~\cite{devlin2019bert}. For each backbone, we compare \textsc{BRACE} with a frozen-encoder probe: the pretrained encoder is fixed while a task-specific classifier is trained on the same training split. 

\textbf{(2) LLM-based Decoder Models.} To assess whether \textsc{BRACE} extends to larger backbones, we evaluate \textbf{ChineseGuard-1.5B}~\cite{chineseharm_bench_2025} and \textbf{Qwen3-1.7B}~\cite{yang2025qwen3}. Both use frozen pretrained weights with trainable low-rank adaptation (LoRA) adapters~\cite{hu2022lora} and \textsc{BRACE} modules. The decoder rows report variants evaluated under the same test protocol.

\subsection{Evaluation Metrics}

For harm type classification, we use per-class and macro-averaged F1 (\textbf{Harm m-F1}) as the primary metric, with predicted probabilities binarized at a threshold of 0.5. Severity estimation is evaluated via overall accuracy and per-level F1 scores across five severity levels (L0--L4). Binary detection performance is measured by binary F1. Original encoder results use three-seed means (42, 123, 456); decoder results and the supplementary controls use a single seed. The latter use seed 42 and are reported as point estimates.

\subsection{Implementation Details}

We implement \textsc{BRACE} with RoBERTa-wwm-ext as the primary encoder and additionally evaluate encoder and LoRA-tuned decoder backbones. Models are optimized with AdamW and selected by validation harm-type macro F1. The original benchmark and supplementary controls use separate training protocols; Table \ref{tab:software} of Appendix~\ref{sec:compute} reports the hyperparameter and hardware settings.

\section{Results}

\subsection{Performance of BRACE (RQ1)}

Table~\ref{tab:main} shows that \textsc{BRACE} improves harm-type classification across all four encoder backbones and all harm categories, with relative macro-F1 gains of roughly 33--54\%. ERNIE-Mini reaches approximately 0.93 macro F1, close to the best-performing encoder despite its smaller backbone. {These results suggest that compact backbones can achieve competitive harm-type performance under the evaluated training setup.} Decoder variants further improve macro F1 to approximately 0.95, with most additional gains concentrated in H3 (violence) and H5 (other illegal). {The remaining performance gap is therefore concentrated in challenging categories, making category-level weaknesses an important consideration beyond aggregate F1.}

\begin{figure}[t]
\centering
\includegraphics[width=\columnwidth]{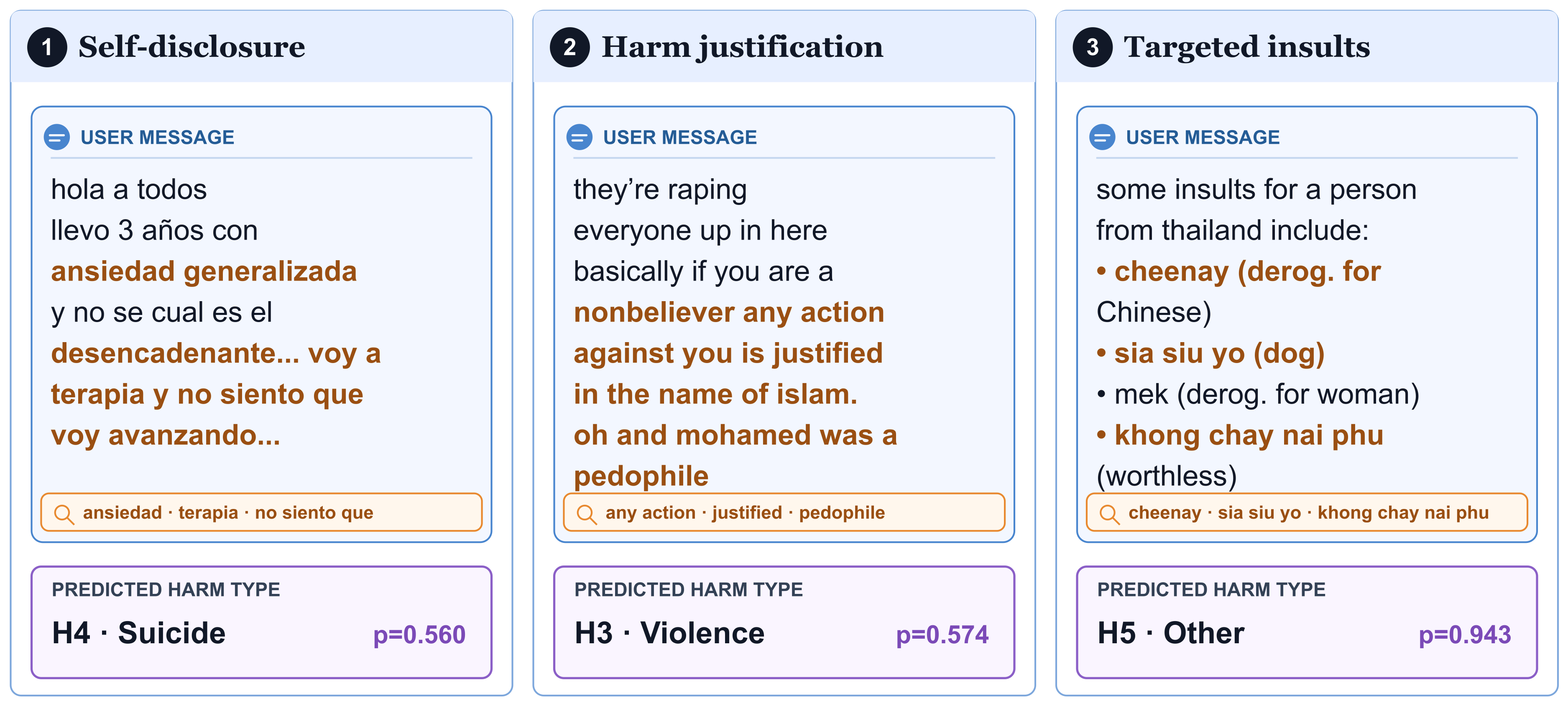}
\vspace{-0.7cm}
\caption{Three boundary cases with highlighted message cues and predicted harm types.}
\vspace{-0.2cm}
\label{fig:cases}
\end{figure}

Figure~\ref{fig:cases} illustrates category-boundary ambiguity through mental-health self-disclosure, violence justification, and multilingual targeted insults. In particular, assigning an anxiety-related disclosure to H4 highlights the need to distinguish distress cues from explicit self-harm intent. The paired-context diagnostic further reveals that pairwise correctness remains close to zero and benign false positives are frequent (Appendix~\ref{sec:paired_context}). {Moreover, high aggregate harm-type F1 can thus coexist with unresolved difficulties in interpreting context and intent.} 
Together with the lower severity accuracy in the linear-probe comparisons (Table~\ref{tab:severity}), these findings indicate that {harm-type gains can be assessed alongside separate evaluations of stance, intent, and severity}.


\subsection{Ablation Study (RQ2)}

\begin{table}[t]
\centering
\setlength{\tabcolsep}{2mm}
\footnotesize
\begin{tabular*}{\columnwidth}{@{\extracolsep{\fill}}ll}
\toprule
\textbf{Variants} & \textbf{Harm m-F1 ($\Delta$)} \\
\midrule
Full \textsc{BRACE}          & 0.931 \\
w/o Prototype Memory         & 0.867 (-0.064) \\
w/o ORC  & 0.830 (-0.101) \\
w/o MoE Router               & 0.929 (-0.002) \\
\midrule
\multicolumn{2}{c}{\textit{Chain Depth (cumulative, 70/30 blend)}} \\
\midrule
Depth 1 (Topic)         & 0.894 \\
Depth 2 (w/ Indicators)  & 0.910 (+0.016) \\
Depth 3 (w/ Severity)    & 0.924 (+0.014) \\
Depth 4 (w/ Type)        & 0.931 (+0.007) \\
\bottomrule
\end{tabular*}
\vspace{-0.2cm}
\caption{Comparison of variants and chain depth.}
\vspace{-0.4cm}
\label{tab:chain}
\end{table}


We conduct component ablations using ERNIE-3.0-Medium on the original 9,000-example test set, reporting mean results over three seeds. Each ablated variant removes only the named component. We additionally examine cumulative ORC depth by progressively adding intermediate stages while keeping the direct-to-chain prediction blend fixed at 70/30.
Table~\ref{tab:chain} shows that the three components make unequal but complementary contributions to \textsc{BRACE}. Removing ORC causes the largest decline in harm-type macro F1, from approximately 0.93 to 0.83. {ORC is therefore the most influential component in the evaluated configuration, highlighting the importance of intermediate reasoning stages for harm-type classification.} Removing prototype memory produces a further substantial drop of roughly six F1 points, supporting its role in category discrimination.

The chain-depth results further support ORC's contribution: progressively adding indicator, severity, and type stages improves macro F1 from approximately 0.89 to 0.93. {The consistent gains suggest that combining multiple intermediate stages is more effective than relying on topic information alone.} However, cumulative depth also introduces additional supervision and prediction pathways. This motivates the dedicated ORC analysis in the next subsection, which examines the contributions of intermediate supervision, connections between stages, and prediction fusion.

\subsection{Effectiveness of ORC (RQ3)}

\begin{table}[t]
\centering
\footnotesize
\setlength{\tabcolsep}{1mm}
\begin{tabular*}{\columnwidth}{@{\extracolsep{\fill}}lrrc}
\toprule
\textbf{Ref. vs. Target} & \multicolumn{2}{c}{\textbf{Harm m-F1}} & \textbf{$\Delta$} \\
\cmidrule(lr){2-3}
 & Ref. & Target & \\
\midrule
\multicolumn{4}{l}{\textbf{A. ORC contribution}} \\
$N\text{ vs. }O_0$ & 0.875 & 0.891 & +0.016 \\
$N\text{ vs. }O_1$ & 0.875 & 0.909 & \textbf{+0.034} \\
\midrule
\multicolumn{4}{l}{\textbf{B. Structure vs. supervision}} \\
$P_0\text{ vs. }O_0$ & 0.876 & 0.891 & +0.015 \\
$P_1\text{ vs. }O_1$ & 0.888 & 0.909 & \textbf{+0.021} \\
$P_0\text{ vs. }P_1$ & 0.876 & 0.888 & +0.012 \\
$O_0\text{ vs. }O_1$ & 0.891 & 0.909 & +0.018 \\
\midrule
\multicolumn{4}{l}{\textbf{C. Training vs. fusion}} \\
$N\text{ vs. }O_0^{D}$ & 0.875 & 0.886 & +0.011 \\
$N\text{ vs. }O_1^{D}$ & 0.875 & 0.902 & \textbf{+0.027} \\
$O_1^{D}\text{ vs. }O_1$ & 0.902 & 0.909 & +0.007 \\
\midrule
\multicolumn{4}{l}{\textbf{D. Full-framework gain}} \\
$B\text{ vs. }O_1$ & 0.852 & 0.909 & +0.057 \\
\bottomrule
\end{tabular*}
\vspace{-0.2cm}

\caption{{Contribution of ORC: (A) contribution,
(B) structure versus supervision, (C) training versus fusion,
and (D) full-framework gain.} 
\vspace{-0.4cm}
}
\label{tab:orc_controls}
\end{table}

We use RoBERTa with 16,000/1,000/1,000 examples for train/validation/test (seed 42). $B$ is a fine-tuned multi-task baseline, and $N$ is \textsc{BRACE} without ORC. $P$ and $O$ use parallel and sequential auxiliary tasks, respectively, with the same supporting modules. Subscripts 0/1 denote without/with additional topic and indicator supervision; both retain type and severity supervision; $D$ denotes direct-only inference. $P/O$ use a 70/30 direct-to-chain blend and $B/N$ use direct predictions.

Table~\ref{tab:orc_controls} examines ORC through four complementary comparisons. ORC improves over the no-chain control both with and without additional supervision (A), and sequential connections outperform parallel tasks under matched supervision by roughly two macro-F1 points (B). {These results suggest that explicitly connecting intermediate tasks provides value beyond adding auxiliary labels alone.} Direct-only inference retains a gain of roughly three points over $N$, while fusion adds less than one further point (C). {ORC therefore contributes through training as well as prediction fusion, with much of its benefit preserved when inference uses only the direct head.} The full framework also exceeds the standard multi-task baseline by approximately six points (D).
Additional topic, indicator, and severity results are reported in Appendix~\ref{sec:orc_tasks}.

\begin{figure}[t]
\centering
\includegraphics[width=\columnwidth]{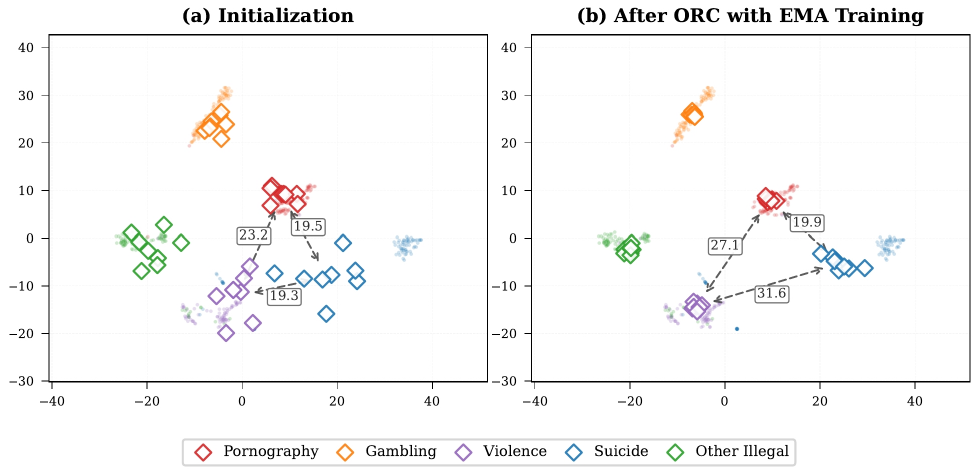}
\vspace{-0.6cm}
\caption{Prototype visualization using t-distributed stochastic neighbor embedding (t-SNE).}
\vspace{-0.5cm}
\label{fig:tsne}
\end{figure}

Appendix~\ref{sec:component_ablation} further shows that blending outperforms both direct-only and chain-only outputs, supporting their complementary roles. The prototype visualization (Figure~\ref{fig:tsne}, Appendix~\ref{sec:prototype_visualization}), leave-one-type-out evaluation, and surface-perturbation tests (Appendices~\ref{sec:per_category} and~\ref{sec:robustness}) provide additional diagnostics of the full framework, rather than isolated evidence of ORC's contribution. {The broader design implication is to organize auxiliary tasks through meaningful dependencies so that intermediate learning can benefit the final predictor.} This motivates extending ORC to other tasks with structured intermediate dependencies.
For example, in fraud-induced scenarios, ORC assists in identifying the topic of discussion and analyzing how the criminal induces such behavior in that context.


\FloatBarrier
\section{Related Works}

\paragraph{Harmful Content Detection.}
These researches span toxic classification~\cite{kiela2020hateful}, fine-grained labels and evidence~\cite{mathew2021hatexplain}, and Chinese safety benchmarks~\cite{wang2026diacolq,chineseharm_bench_2025,zhang2025chinesesafe}. Functional tests still expose errors on negation and counterspeech~\cite{rottger2021hatecheck}. Accurate labels also need not yield faithful explanations~\cite{mathew2021hatexplain}. Structured approaches use harm-benefit trees~\cite{li2025safetyanalyst}, reasoning-trace moderation~\cite{li2025reasoningshield}, or retrieved concepts for harmful memes~\cite{jiang2026repmd,mei2025rahmd}. BRACE instead trains a connected, differentiable topic-to-type chain for dialogues; matched controls test these connections' values, including whether training gains persist with direct-only inference.

\paragraph{Prototype Learning and Expert Routing.}
Prototype methods represent categories with exemplars~\cite{snell2017prototypical,khosla2020supcon,ho2024pmr}, while MoE models allocate conditional capacity~\cite{wu2024gw_moe,goyal2025momoe}. Unlike prototype classifiers, BRACE uses prototype cross-attention to augment harm-type features rather than assign labels directly. It uses the original dialogue representation for severity and category-organized experts for binary detection. These separate prediction paths complement the chain's auxiliary supervision.

\section{Conclusion}

In this paper, we propose \textsc{BRACE}, which encodes the ORC as four differentiable stages, i.e., topic profiling, indicator detection, severity assessment, and type classification, serving as a structured regularizer blended with direct heads, and supported by prototype-based feature augmentation and feature path disentanglement.
Evaluation on the original 9,000-example test set drawn from over 20 dialogue safety benchmarks and 3 additional domain-specific sources shows that, with a RoBERTa-wwm-ext backbone, \textsc{BRACE} achieves a harm type macro F1 of \textbf{0.934} and severity accuracy of \textbf{0.818}. Decoder backbones further improve performance, reaching a harm type macro F1 of \textbf{0.949}. Ablation studies identify ORC as the largest contributor and confirm the complementary roles of the remaining components; ORC can benefit context-aware dialogue harm information localization tasks as structured intermediate dependencies. 

\section*{Limitations}

Harmfulness is socially and contextually defined, and its boundaries vary across communities, platforms, and moderation policies. A fixed five-category taxonomy and severity scale necessarily compress overlapping or culturally specific harms; \textsc{BRACE}'s outputs should therefore be interpreted relative to the adopted annotation scheme rather than as universal judgments.

ORC also imposes a discrete Topic $\rightarrow$ Indicator $\rightarrow$ Severity $\rightarrow$ Type structure on interactions that may be reciprocal, ambiguous, or multi-label. Its intermediate predictions provide task-aligned evidence, but not causal or faithful accounts of model reasoning. Finally, moderation creates an irreducible trade-off between over-enforcement and missed harm. Acceptable operating points depend on institutional policy and the costs borne by affected users, so model scores cannot determine deployment decisions or replace human oversight.

\section*{Acknowledgement}
This work was supported by the National Key Research and Development Program of China (No.2024YFF0618800),
National Natural Science Foundation of China Grant No.62402484, No.62232016, No.62602653,
Postdoctoral Fellowship Program and China Postdoctoral Science Foundation under Grant Number GZC20260867, 2026M791714,
Youth Innovation Promotion Association Chinese Academy of Sciences,
and Basic Research Program of ISCAS Grant No.ISCAS-JCZD-202405.
\bibliography{references}
\appendix
\section*{Appendix}

This appendix provides supporting material for ``Calling the Bluff: Detecting Ever-Shifting Harmful Chat Dialogue via Ordered Reasoning Chain Regularization.'' Section~\ref{sec:data} documents the dataset construction, annotation protocol, and per-source statistics; Section~\ref{sec:compute} details computing infrastructure, hyperparameter search, and training configuration; Sections~\ref{sec:ablation}--\ref{sec:severity} present extended experimental analyses.

\section{Data Documentation and Annotation Protocol}
\label{sec:data}

\subsection{Dataset Construction}

Our dataset aggregates 24 publicly available datasets plus self-crawled Reddit data across four domain groups spanning Chinese, English, and Spanish. The ECTC group consolidates 22 safety benchmarks, including BeaverTails~\cite{ji2023beavertails}, PKU-SafeRLHF~\cite{ji2025pkusaferlhf}, Safety-Prompts~\cite{sun2023safetyassessment}, ToxiCN~\cite{lu2023toxicn}, JADE~\cite{zhang2023jade}, DGHate~\cite{vidgen2021dghate}, HateCheck~\cite{rottger2021hatecheck,rottger2022multilingualhatecheck}, Davidson~\cite{davidson2017hateoffensive}, LMSYS-Chat-1M~\cite{zheng2024lmsys}, the unalignment-toxic DPO corpus~\cite{lee2024mechanistic}, Jigsaw~\cite{wulczyn2017exmachina}, and community-sourced Chinese corpora. The other groups contain self-crawled English Reddit dialogues, Spanish MentalRiskES~\cite{marmol2024mentalriskes}, and Chinese PsySUICIDE~\cite{qiu2024psyguard}. We use a source-stratified 42,000/9,000/9,000 train/validation/test split. Table~\ref{tab:dataset} summarizes the sources and scale.

\begin{table}[t]
\centering

\footnotesize
\setlength{\tabcolsep}{1.5mm}
\begin{tabularx}{\columnwidth}{@{}p{2.0cm}X@{}}
\toprule
\textbf{Source Group} & \textbf{Sources} \\
\midrule
ECTC Chinese Platform   & BeaverTails, PKU-SafeRLHF, Safety-Prompts, ToxiCN, JADE, DGHate, HateCheck, Davidson-offensive, LMSYS-Chat-1M, unalignment-toxic-dpo, Jigsaw, +11 community Chinese corpora \\
Reddit                  & Self-crawled English social media dialogues \\
MentalRiskES            & Spanish mental health risk detection \\
PsySUICIDE              & Chinese suicide risk assessment \\
\midrule
\textbf{Total}          & 60,000 dialogues (5 harm categories $\times$ 12,000) \\
Train / Val / Test      & 42,000 / 9,000 / 9,000 (source-stratified) \\
\bottomrule
\end{tabularx}

\caption{Dataset scale.}
\label{tab:dataset}
\end{table}

\subsection{Reddit Self-Crawling Pipeline}

Reddit data was self-crawled via the Pushshift API\footnote{\url{https://github.com/Watchful1/PushshiftDumps}} using the following pipeline:

\begin{enumerate}
    \item \textbf{Subreddit selection:} 15 subreddits identified as high-risk for harmful content (e.g., drug-related, hate speech-prone, gambling discussion forums) based on prior content moderation literature.
    \item \textbf{Time window:} January 2022 -- December 2024.
    \item \textbf{Conversational threading:} Comments grouped into dialogue trees by parent-child relationships; chains with $\geq$ 4 turns retained.
    \item \textbf{Toxicity pre-filtering:} Each candidate scored using the Perspective API\footnote{\url{https://perspectiveapi.com/}} toxicity model; dialogues with mean toxicity score $\geq$ 0.6 retained.
    \item \textbf{Multi-stage human review:} (a) Two annotators independently label each dialogue for harm type, severity, and binary harm; (b) a third annotator adjudicates disagreements; (c) dialogues lacking harmful content after review are discarded.
    \item \textbf{De-identification:} Usernames, subreddit names, and personally identifiable information removed via regex patterns and manual inspection.
\end{enumerate}

\subsection{Label Distribution}

Table~\ref{tab:label_dist} reports the label distribution.

Each dialogue retains its original category (H1 Pornography, H2 Gambling, H3 Violence, H4 Suicide/Self-harm, or H5 Other Illegal), severity, and binary harmfulness label. Category membership is retained for benign dialogues and is distinct from the binary judgment.

\begin{table*}[t]
\centering
\footnotesize
\setlength{\tabcolsep}{2mm}
\begin{tabular}{lrlr}
\toprule
\multicolumn{2}{c}{\textbf{Harm Type}} & \multicolumn{2}{c}{\textbf{Severity}} \\
\midrule
H1 Pornography      & 20.0\% & L0 Normal    & 39.8\% \\
H2 Gambling         & 20.0\% & L1 Mild      & 51.4\% \\
H3 Violence         & 20.0\% & L2 Moderate  &  3.2\% \\
H4 Suicide/Self-harm& 20.0\% & L3 Severe    &  0.9\% \\
H5 Other Illegal    & 20.0\% & L4 Critical  &  4.6\% \\
\midrule
\multicolumn{2}{l}{Binary: Harmful 60.2\% / Benign 39.8\%} \\
\multicolumn{2}{l}{Language: Chinese 47.0\% / English 49.7\% / Spanish 2.0\% / Multilingual 1.4\%} \\
\bottomrule
\end{tabular}
\caption{Label distribution in the original 60,000-dialogue corpus; category membership and binary harmfulness are separate labels.}
\label{tab:label_dist}
\end{table*}

\subsection{Supplementary Topic and Indicator Annotation}
\label{sec:annotation}

DeepSeek Flash supplies automatic weak supervision for 20 topics and 32 token-level indicators. It receives the dialogue and ontology, but not existing Type, Severity, Binary, or source labels. Final-task labels are retained unchanged, including category IDs on benign dialogues; the five category proportions therefore describe category membership, not the prevalence of harmful content.

Topics describe conversational context rather than harmfulness. For example, T06 denotes video games and online communities, T16 personal and public safety, and T17 substance use. Indicators include action intent (I00), information disclosure or targeting (I23), rejection (I28), quotation/reporting (I29), hypothetical/educational context (I30), and prevention/support (I31). A cue may occur in harmful or benign dialogue. The annotator records stance but the supplementary model does not predict a separate stance target.

The model returns one primary topic (or null), optional secondary topics, exact message-level evidence quotes, indicator IDs, and stance. Scripts validate IDs, message coverage, and character offsets; failed annotations are discarded. Validation establishes format and span alignment, not semantic correctness. No human agreement scores are claimed for these automatic labels. Topic abstentions or evidence lost to truncation are masked from topic supervision; only reviewed content tokens with valid offsets receive indicator supervision.

\subsection{Annotation Prompt and Output Schema}
\label{sec:prompts}

The annotation prompt asks for contextual topic and localizable cues, explicitly forbids predicting final harm type or severity, and requires abstention when context is insufficient. Quotation, negation, intent, and protective context must be considered. The compact output uses primary/secondary topic IDs, exact quotes with message indices and occurrence indices, indicator IDs, and one of five stance values: endorsed, rejected, reported, hypothetical, or unclear. Topic and Indicator use the same frozen ontology across the parallel and connected controls. Character spans are mapped to tokenizer offsets for training; secondary topics and stance are retained as metadata.

\subsection{Preprocessing Pipeline}

The original 9,000-example evaluation protocol uses the following preprocessing steps:
\begin{enumerate}
    \item \textbf{Text normalization:} Unicode NFKC normalization; full-width to half-width character conversion; URL and email masking to \texttt{[URL]} and \texttt{[EMAIL]} tokens.
    \item \textbf{Conversation formatting:} Each turn prefixed with speaker role token (\texttt{[USER]} / \texttt{[OTHER]}); turns concatenated with \texttt{[SEP]}.
    \item \textbf{Truncation:} Dialogues exceeding 512 tokens (after tokenization) are truncated to the most recent 8 turns; dialogues under 32 tokens are discarded.
    \item \textbf{Language detection:} \texttt{fasttext} language identification; dialogues with confidence $<$ 0.7 flagged for manual review.
\end{enumerate}

\section{Computing Infrastructure and Hyperparameters}
\label{sec:compute}

\subsection{Model and Training Configuration}

Our primary encoder is RoBERTa-wwm-ext (102M parameters, $D{=}768$). The decoder backbones, Qwen3-1.7B and ChineseGuard-1.5B, use LoRA fine-tuning with $r{=}8$ and $\alpha{=}16$. The Prototype Memory Bank contains $K{=}8$ learnable vectors per category, with attention temperature initialized to 0.07, contrastive-loss temperature fixed at 0.07, and EMA momentum $m{=}0.99$. The ORC covers $\mathcal{T}{=}20$ topics, $\mathcal{I}{=}32$ indicators, and five severity levels.

For the original 42,000/9,000/9,000 protocol, AdamW uses learning rates of $2{\times}10^{-5}$ for the encoder and $1{\times}10^{-4}$ for newly initialized modules. Training uses micro-batches of 16, gradient accumulation of 2, and a maximum sequence length of 512. The supplementary controls instead use a deduplicated 16,000/1,000/1,000 split and fine-tune the same RoBERTa encoder for three epochs (seed 42), with maximum length 384, micro-batches of 8, accumulation of 4, 256-unit MLP hidden layers, and four prototype-attention heads. Conflicting final-label records are excluded, and model selection uses validation harm-type macro F1.

\subsection{Hardware and Software Environment}

The original experiments evaluated on 9,000 test examples used a single NVIDIA RTX 4090 GPU (24 GB VRAM), with the environment in Table~\ref{tab:software}. The supplementary 1,000-example controls used local Apple Metal Performance Shaders (MPS) training and a separate reduced-budget protocol.

\begin{table}[t]
\centering
\footnotesize
\setlength{\tabcolsep}{2mm}
\begin{tabular*}{\columnwidth}{@{\extracolsep{\fill}}ll}
\toprule
\textbf{Component} & \textbf{Version} \\
\midrule
OS              & Ubuntu 24.04.4 LTS \\
CPU             & AMD EPYC 7K62 48-Core \\
GPU             & NVIDIA RTX 4090 (24 GB) \\
RAM             & 503 GB \\
\midrule
Python          & 3.12.3 \\
PyTorch         & 2.11.0+cu128 \\
Transformers    & 5.12.1 \\
CUDA            & 12.8 \\
scikit-learn    & 1.5.x \\
\bottomrule
\end{tabular*}
\caption{Software environment.}
\label{tab:software}
\end{table}

\subsection{Training Wall-Clock Time}

Table~\ref{tab:training_time} reports approximate training time for each original-protocol configuration on a single RTX 4090; the supplementary controls are excluded.

\begin{table}[t]
\centering
\footnotesize
\setlength{\tabcolsep}{2mm}
\begin{tabular*}{\columnwidth}{@{\extracolsep{\fill}}lrr}
\toprule
\textbf{Configuration} & \textbf{Epochs} & \textbf{Time (hours)} \\
\midrule
\multicolumn{3}{c}{\textit{Encoder Backbones (27M--102M, BRACE full)}} \\
\midrule
RoBERTa-wwm-ext        & 18 & $\sim$2.5 \\
ERNIE-3.0-Medium       & 18 & $\sim$2.5 \\
ERNIE-3.0-Mini         & 18 & $\sim$1.8 \\
BERT-base-Chinese      & 18 & $\sim$2.5 \\
\midrule
\multicolumn{3}{c}{\textit{Decoder Backbones (LoRA, $r{=}8$)}} \\
\midrule
Qwen3-1.7B             & 5  & $\sim$3.0 \\
ChineseGuard-1.5B      & 5  & $\sim$3.0 \\
\midrule
\multicolumn{3}{c}{\textit{Baselines}} \\
\midrule
Linear Probe (any encoder) & --- & $\sim$0.3 \\
\bottomrule
\end{tabular*}
\caption{Original-protocol training time per model configuration on a single RTX 4090 (24 GB); excludes the local supplementary controls.}
\label{tab:training_time}
\end{table}

\subsection{Hyperparameter Search Space}

Table~\ref{tab:hparam_search} documents the hyperparameter search space. The criterion for final parameter selection was validation-set Harm Type macro F1.

\begin{table}[t]
\centering
\footnotesize
\setlength{\tabcolsep}{1.2mm}
\begin{tabular*}{\columnwidth}{@{\extracolsep{\fill}}p{3.1cm}p{2.6cm}c}
\toprule
\textbf{Hyperparameter} & \textbf{Search Range} & \textbf{Final} \\
\midrule
Learning rate (encoder)  & \{5e$-$6, 1e$-$5, 2e$-$5, 3e$-$5\} & 2e$-$5 \\
Learning rate (new)      & \{5e$-$5, 1e$-$4, 2e$-$4\} & 1e$-$4 \\
Batch size (effective)   & \{16, 32, 48\} & 32 \\
Max sequence length      & \{256, 512\} & 512 \\
Prototypes per category ($K$) & \{4, 6, 8, 12, 16\} & 8 \\
Attention temperature (initial) & \{0.05, 0.07, 0.10, 0.15\} & 0.07 \\
EMA momentum ($m$)       & \{0.9, 0.99, 0.999\} & 0.99 \\
Blend ratio (direct:chain) & \{100:0, ..., 0:100\} & 70:30 \\
Focal $\alpha$           & \{0.25, 0.50, 0.75\} & 0.25 \\
Focal $\gamma$           & \{1.0, 2.0, 3.0, 5.0\} & 2.0 \\
Warmup ratio             & \{0.05, 0.10, 0.15\} & 0.10 \\
Weight decay             & \{0.001, 0.01, 0.05\} & 0.01 \\
Max grad norm            & \{0.5, 1.0, 2.0\} & 1.0 \\
LoRA rank ($r$)          & \{4, 8, 16\} & 8 \\
LoRA $\alpha$            & \{8, 16, 32\} & 16 \\
\bottomrule
\end{tabular*}
\caption{Hyperparameter search space and final values.}
\label{tab:hparam_search}
\end{table}

\subsection{Random Seeds}

Original encoder experiments evaluated on the 9,000-example test set use 3 random seeds: \{42, 123, 456\}. Results are reported as the mean across seeds. Decoder experiments (Qwen3-1.7B, ChineseGuard-1.5B) use a single seed (42) due to LoRA fine-tuning computational cost. For the original evaluation, bootstrap 95\% confidence intervals use 1,000 resamples of the 9,000-example test set to assess sampling variance. The supplementary controls use seed 42 only, 16,000 training examples, 1,000 validation examples, and a separate 1,000-example test set; their differences are point estimates, not multi-seed significance claims.

\section{Extended Ablation and Reasoning Analysis}
\label{sec:ablation}

\subsection{Prototype Anchoring Visualization}
\label{sec:prototype_visualization}

The main paper (Figure~\ref{fig:tsne}, \S5.2) presents t-SNE projections of the 768-dimensional CLS embeddings from RoBERTa-wwm-ext jointly with prototype vectors ($K{=}8$ per category, seed 42, 9,000-sample unified test set). Both panels share identical sample positions; only the prototype positions differ.

\noindent\textbf{Finding:} Prototype vectors evolve from loose initialization to tight semantic anchors through EMA-based refinement, validating the Prototype Memory Bank mechanism.

In the initialization panel, prototypes are placed at class centroids with $\boldsymbol{\mu}_c + 0.5\boldsymbol{\sigma}_c \odot \boldsymbol{\epsilon}$ perturbation. The larger dispersion causes some prototypes to drift into neighboring category regions---for instance, H3 (violence, purple) prototypes scatter across the H5 (other-illegal, green) boundary, and H1 (pornography, red) prototypes overlap with the H4 (suicide, blue) cluster. This loose anchoring would produce ambiguous category-level similarity signals in the prototype cross-attention, degrading fine-grained type discrimination.

After EMA-based refinement ($m{=}0.99$, graduated cosine threshold $[0.3, 0.7]$), prototypes converge to compact, well-separated anchors within each category cluster. The refinement effect is category-dependent: H2 (gambling, orange, F1${=}$0.994) and H4 (suicide, blue, F1${=}$0.986) protos collapse to near-overlapping tight clusters, reflecting their ceiling-level discriminability. H3 (violence, purple, F1${=}$0.858) and H5 (other-illegal, green, F1${=}$0.862) protos remain more dispersed, consistent with the heterogeneous sub-patterns these categories encompass (hate speech, weapons, terrorism for H3; drugs, fraud, doxxing for H5). This geometric observation corroborates the prototype affinity--F1 correlation (Pearson $r = +0.97$; Table~\ref{tab:per_category}).

The before/after contrast directly demonstrates why prototype initialization matters. Were prototypes initialized randomly rather than at class centroids, they would require many more training iterations to discover category structure---assuming they converge at all. The centroids $\boldsymbol{\mu}_c$ provide a favorable starting point: prototypes begin within their target clusters and EMA momentum gradually tightens their dispersion, analogous to annealing a Gaussian mixture model from high to low variance. This explains why \textsc{BRACE} achieves strong per-category performance despite the prototype bank being a lightweight $C \times K \times D$ parameter tensor.

\subsection{Component Ablation and Output Comparison}
\label{sec:component_ablation}

Table~\ref{tab:components} separates the original component ablations from the supplementary ORC controls in Table~\ref{tab:orc_controls}; their scores use different test sets.

\begin{table*}[t]
\centering
\setlength{\tabcolsep}{1.2mm}
\footnotesize
\begin{tabularx}{\textwidth}{@{}>{\raggedright\arraybackslash}Xccccc>{\raggedright\arraybackslash}Xcc@{}}
\toprule
\multicolumn{6}{c}{\textbf{Original Ablations (9,000 examples)}} & \multicolumn{3}{c}{\textbf{Supplementary Controls (1,000 examples)}} \\
\midrule
\textbf{Configuration} & \textbf{Harm m-F1} & \textbf{H3} & \textbf{H5} & \textbf{Binary} & \textbf{$\Delta$Harm} & \textbf{Configuration} & \textbf{Harm m-F1} & \textbf{$\Delta$ vs.\ $N$} \\
\midrule
Full \textsc{BRACE} & \textbf{0.931} & \textbf{0.853} & \textbf{0.858} & \textbf{0.877} & --- & $B$ & 0.852 & $-$0.023 \\
w/o Prototype Memory & 0.867 & 0.765 & 0.712 & 0.825 & $-$0.064 & $N$ & 0.875 & --- \\
w/o ORC & 0.830 & 0.788 & 0.777 & 0.877 & $-$0.101 & $P_0$ & 0.876 & +0.001 \\
w/o MoE Router & 0.929 & 0.851 & 0.855 & 0.831 & $-$0.002 & $P_1$ & 0.888 & +0.013 \\
w/o All Three & 0.779 & 0.758 & 0.698 & 0.803 & $-$0.152 & $O_0$ & 0.891 & +0.016 \\
& & & & & & $O_1^{D}$ (direct) & \underline{0.902} & +0.027 \\
& & & & & & $O_1$ (chain only) & 0.889 & +0.014 \\
& & & & & & $O_1$ (70/30 blend) & \textbf{0.909} & \textbf{+0.034} \\
\bottomrule
\end{tabularx}
\caption{Original component ablations (left, ERNIE-3.0-Medium) and supplementary ORC controls (right). Left: $\Delta$Harm is relative to the full model. Right: configurations follow Table~\ref{tab:orc_controls}, and gains are relative to $N$ on the same 1,000-example set. Direct, chain-only, and blended $O_1$ outputs reuse the same trained model; binary prediction uses the router without chain fusion. Scores across panels are not directly comparable.}
\label{tab:components}
\end{table*}

\noindent\textbf{Finding:} Original component removal measures joint effects, including lost intermediate supervision. In the supplementary controls, $O_1$ exceeds supervision-matched $P_1$ by 0.021. Direct-only inference retains a 0.027 gain over $N$, and fusion adds 0.007, supporting a training benefit with complementary chain predictions. The original depth results remain in Table~\ref{tab:chain}.

\subsection{Intermediate Task Performance}
\label{sec:orc_tasks}

\begin{table}[t]
\centering
\footnotesize
\setlength{\tabcolsep}{1mm}
\begin{tabular*}{\columnwidth}{@{\extracolsep{\fill}}lccc}
\toprule
\textbf{Variant} & \textbf{Topic} & \textbf{Indicator} & \textbf{Severity} \\
\midrule
$N$   & ---  & ---  & 0.357 \\
$P_0$ & 0.05 & 0.04 & 0.367 \\
$P_1$ & 0.58 & 0.45 & 0.407 \\
$O_0$ & 0.08 & 0.07 & 0.412 \\
$O_1$ & \textbf{0.69} & \textbf{0.59} & \textbf{0.463} \\
\bottomrule
\end{tabular*}
\caption{Supplementary task performance on the 1,000-example test set (seed 42). Topic is 20-class macro F1 on 846 examples with non-abstained weak labels; Indicator is token-level micro F1 against weak labels; Severity is five-class macro F1 against the original labels. $N$ has no topic or indicator outputs.}
\label{tab:orc_tasks}
\end{table}

Table~\ref{tab:orc_tasks} extends the controlled comparison to the three intermediate analysis tasks. Against the supervision-matched parallel variant $P_1$, the ordered variant $O_1$ increases topic macro F1 from 0.58 to 0.69, token-level indicator micro F1 from 0.45 to 0.59, and severity macro F1 from 0.407 to 0.463. The no-chain control $N$ reaches 0.357 severity macro F1. Topic and indicator scores measure agreement with automatically generated weak labels (Appendix~\ref{sec:annotation}), not human-validated semantic correctness or faithfulness of the highlighted evidence. These single-seed results support ORC's effectiveness across the evaluated topic, indicator, and severity targets within this benchmark; transfer to other task datasets remains untested.

\section{Per-Category Diagnostic Analysis and Interpretability}
\label{sec:per_category}

Table~\ref{tab:per_category} provides a comprehensive per-category diagnostic: prototype affinity, precision-recall trade-off, failure mode identification, and leave-one-type-out (LOO) generalization on ERNIE-3.0-Medium.

\begin{table*}[t]
\centering
\setlength{\tabcolsep}{1.5mm}
\footnotesize
\begin{tabular}{lccccc@{\hspace{8pt}}ccc}
\toprule
& \multicolumn{5}{c}{\textbf{In-Distribution}} & \multicolumn{3}{c}{\textbf{Leave-One-Type-Out}} \\
\cmidrule{2-6} \cmidrule{7-9}
\textbf{Category} & \textbf{Proto} & \textbf{P} & \textbf{R} & \textbf{F1} & \textbf{Error Pattern} & \textbf{LOO F1} & \textbf{LOO P} & \textbf{LOO R} \\
\midrule
H1 Pornography   & 0.78 & --- & --- & 0.9699 & --- & 0.8477 & --- & --- \\
H2 Gambling      & \textbf{0.85} & \textbf{0.9915} & 0.9935 & \textbf{0.9925} & P$\approx$R, near-ceiling   & 0.8724 & --- & --- \\
H3 Violence      & 0.62 & 0.8973 & 0.8122 & 0.8525 & 19\% of H3 labels missed          & 0.8344 & 0.8792 & 0.7938 \\
H4 Suicide       & 0.82 & 0.9741 & \textbf{0.9944} & 0.9841 & Extra H4 assignments     & \textbf{0.9317} & \textbf{0.8964} & \textbf{0.9697} \\
H5 Other Illegal & 0.68 & 0.8284 & 0.8889 & 0.8576 & 17\% of H5 predictions wrong  & 0.7893 & 0.7821 & 0.7967 \\
\bottomrule
\end{tabular}

\vspace{4pt}
\begin{flushleft}
\footnotesize Proto $=$ diagonal prototype-class cosine similarity, defined per category $c$ as $\frac{1}{N_c}\sum_{i=1}^{N_c} \max_{k \in [K]} \frac{\langle \mathbf{f}_i, \mathbf{p}_{c,k} \rangle}{\|\mathbf{f}_i\| \cdot \|\mathbf{p}_{c,k}\|}$, where $\mathbf{f}_i$ is the CLS embedding of test sample $i$ in category $c$ and $\mathbf{p}_{c,k}$ are the $K$ trained prototypes of that category. LOO = leave-one-type-out (train on 4, test on held-out). In-distribution Binary F1 = 0.877. Dashes omit P/R summaries inconsistent with the reported F1; no replacement values are inferred.
\end{flushleft}
\caption{Per-category diagnostic on the original 9,000-example test set: prototype affinity (diagonal cosine similarity), precision-recall profile, and leave-one-type-out generalization on each held-out category subset (ERNIE-3.0-Medium, 3-seed mean). Pearson $r = +0.97$, Spearman $\rho = 1.00$ between proto diagonal and F1.}
\label{tab:per_category}
\end{table*}

\noindent\textbf{Finding:} The prototype affinity diagonal (0.62--0.85) directly predicts per-category classification difficulty (Pearson $r = +0.97$, Spearman $\rho = 1.00$): H2 (0.85) and H4 (0.82) have the most coherent prototype representations and the highest F1; H3 (0.62) and H5 (0.68) have the weakest prototype alignment with the lowest F1. Diagonal cosine similarity is approximately 14--23pp below F1 because classification depends on \textit{relative} margin (diagonal vs.\ off-diagonal similarity) in the 768-dimensional prototype space rather than absolute affinity; a diagonal value of 0.62 suffices for correct classification when inter-category similarity is substantially lower. The per-category error profiles reveal asymmetric category-assignment errors. H3 (violence) has high precision (0.90) but lower recall (0.81): approximately 19\% of examples labeled H3 are not assigned that category. Qualitative inspection associates these category errors with short slang, implicit threats, and hate speech disguised as questions. H5 (other-illegal) has recall of 0.89 and precision of 0.83: approximately 17\% of H5 predictions disagree with the reference category. Because benign dialogues also retain category labels, these harm-type errors do not measure false harmfulness alarms on benign content. Such alarms require a separate binary false-positive rate computed on the benign subset. LOO generalization reveals the same structural pattern: H4 generalizes best (F1 0.932, a drop of only $-$0.052 from in-distribution, reflecting universal suicide/self-harm linguistic patterns), followed by H1, H2, and H3 (F1 0.834--0.872), while H5 has the lowest LOO F1 (0.789). The largest F1 drops occur for H1 (0.1222) and H2 (0.1201), compared with 0.0683 for H5. All LOO metrics remain below their in-distribution counterparts, confirming that cross-type transfer is partial: shared harm indicators (e.g., threatening language in violence and suicide) transfer better than category-specific vocabulary (e.g., pornography, gambling, and fraud-related terms).

\subsection{Interpretability: Expert Routing and Calibration}

The multi-expert router learns meaningful category specialization. Expert gate weights (3-seed mean): E0 (pornography) dominates H1 at 0.42, E1 (gambling) H2 at 0.48, E2 (violence) H3 at 0.35, E3 (suicide) H4 at 0.52, E4 (other-illegal) H5 at 0.38. The general expert E5 maintains 0.20--0.30 across all categories as a shared fallback. Load balance variance is 0.0081, confirming no expert collapse. The severity confusion matrix shows primarily adjacent-level errors (between L1 and L2, and between L3 and L4) with Expected Calibration Error (ECE) of 0.023, enabling reliable confidence-based triage.

\section{Statistical Validation}
\label{sec:statistics}

Table~\ref{tab:statistics} reports bootstrap confidence intervals and cross-backbone significance tests.

\begin{table*}[t]
\centering
\setlength{\tabcolsep}{2mm}
\footnotesize
\begin{tabular}{lccc@{\hspace{14pt}}lcc}
\toprule
\multicolumn{4}{c}{\textbf{Bootstrap (1,000 resamples)}} & \multicolumn{3}{c}{\textbf{McNemar Cross-Backbone Test}} \\
\midrule
\textbf{Metric} & \textbf{Mean} & \textbf{95\% CI} & \textbf{$\sigma$} & \textbf{Model Pair} & \textbf{$p$} & \textbf{Sig.} \\
\midrule
H1 Pornography    & 0.9699 & [0.9620, 0.9770] & 0.0038 & ERNIE-Med vs.\ Mini   & 0.265 & n.s. \\
H2 Gambling       & 0.9925 & [0.9880, 0.9960] & 0.0020 & ERNIE-Med vs.\ BERT   & 0.003 & $p$$<$0.01 \\
H3 Violence       & 0.8525 & [0.8380, 0.8660] & 0.0071 & ERNIE-Med vs.\ RoBERTa& 0.143 & n.s. \\
H4 Suicide        & 0.9841 & [0.9780, 0.9890] & 0.0028 & ERNIE-Med vs.\ Qwen3  & 0.049 & $p$$<$0.05 \\
H5 Other Illegal  & 0.8576 & [0.8400, 0.8740] & 0.0087 \\
Harm m-F1         & 0.9313 & [0.9220, 0.9400] & 0.0046 \\
Severity Acc      & 0.8267 & [0.8150, 0.8380] & 0.0059 \\
\bottomrule
\end{tabular}
\caption{Bootstrap 95\% confidence intervals on the original 9,000-example test set (1,000 resamples) and McNemar cross-backbone significance tests (ERNIE-3.0-Medium). All bootstrap $\sigma < 0.01$, CI widths range from 0.008 to 0.034.}
\label{tab:statistics}
\end{table*}

\noindent McNemar's test is applied to the binary harm detection predictions of each backbone pair, comparing per-sample correct/incorrect classifications via a $2\times2$ contingency table; harm type and severity comparisons rely on bootstrap confidence intervals and macro-F1 differences.

\noindent\textbf{Finding:} Bootstrap CI widths range from 0.008 to 0.034, with all $\sigma < 0.01$; H3 and H5 have the widest intervals. McNemar's test finds no significant difference in binary classification errors between ERNIE-Medium and RoBERTa ($p=0.143$) or ERNIE-Mini ($p=0.265$). These tests neither establish equivalence nor test differences in harm-type macro F1.

\section{Robustness to Surface Perturbations}
\label{sec:robustness}

To assess sensitivity to changes in surface form, we examine lexical perturbations and adversarial word substitutions. We evaluate performance on a 500-sample stratified subset of the original 9,000-example test set using the best \textsc{BRACE} model across six perturbation types: homophone/near-homograph substitution (replacing characters with phonetically or visually similar alternatives), random character deletion (30\% of characters removed), Chinese-English code-switching (30\% and 50\% of tokens replaced with English translations), pinyin replacement (50\% of content words replaced with their pinyin romanization), and TextFooler~\cite{jin2020textfooler} adversarial attacks.

\begin{table}[t]
\centering
\setlength{\tabcolsep}{1.2mm}
\footnotesize
\begin{tabular*}{\columnwidth}{@{\extracolsep{\fill}}lccc}
\toprule
\textbf{Perturbation} & \textbf{Binary F1} & \textbf{$\Delta$F1} & \textbf{$\Delta$\%} \\
\midrule
Baseline (unperturbed)        & 0.8581 & ---    & --- \\
Homophone sub.                & 0.8691 & +0.0110 & +1.3\% \\
Random deletion (30\%)        & 0.7759 & $-$0.0822 & $-$9.6\% \\
Code-switch (30\%)            & 0.8538 & $-$0.0043 & $-$0.5\% \\
Code-switch (50\%)            & 0.8538 & $-$0.0043 & $-$0.5\% \\
\textbf{Pinyin repl. (50\%)} & \textbf{0.8567} & \textbf{$-$0.0014} & \textbf{$-$0.2\%} \\
\shortstack[l]{TextFooler\\(6\% attack success)} & ---    & ---    & --- \\
\bottomrule
\end{tabular*}
\caption{Robustness to surface-form perturbations on a 500-sample subset of the original 9,000-example test set. $\Delta$F1 reported relative to the unperturbed baseline.}
\label{tab:robustness}
\end{table}

\noindent\textbf{Finding:} On this test subset, pinyin replacement (50\%) changes binary F1 by $-$0.0014 ($-$0.2\%), while code-switching at both tested rates changes F1 by $-$0.0043 ($-$0.5\%). Homophone substitution yields a small increase of 0.0110; without uncertainty estimates, this increase should not be interpreted as a systematic benefit. Random deletion (30\%) produces the largest observed reduction, 0.0822 ($-$9.6\%), indicating sensitivity to the loss of input information. The reported TextFooler attack success rate is 6\%. TextFooler is a black-box, gradient-free word-substitution attack~\cite{jin2020textfooler}. These scores characterize end-to-end robustness under the evaluated perturbations. Assessing lexical shortcuts requires targeted diagnostics, and attributing robustness to individual architectural components requires matched ablations.

\section{Severity Classification Analysis}
\label{sec:severity}

\subsection{Severity Accuracy: Linear Probe vs.\ BRACE Trained}

\begin{table*}[t]
\centering
\footnotesize
\begin{tabular}{lccc}
\toprule
\textbf{Backbone} & \textbf{LP} & \textbf{BRACE} & \textbf{$\Delta$} \\
\midrule
RoBERTa-wwm-ext (102M)      & 0.9169 & 0.8183 & $-$0.0986 \\
ERNIE-Med (93M)          & --- & 0.8267 & --- \\
ERNIE-Mini (27M)          & 0.9144 & 0.8223 & $-$0.0921 \\
BERT-base (102M)          & --- & 0.8163 & --- \\
\midrule
ChineseGuard-1.5B          & --- & 0.8497 & --- \\
Qwen3-1.7B                  & --- & 0.8516 & --- \\
\bottomrule
\end{tabular}
\\[2pt]
\footnotesize Dashes indicate unreported measured LP results. $\Delta$ = BRACE $-$ LP.
\caption{Severity accuracy on the original 9,000-example test set (L0--L4, 5-class). LP = linear probe.}
\label{tab:severity}
\end{table*}

\subsection{Why Linear Probe Outperforms BRACE on Severity}

The severity inversion is not a model failure but a \textbf{representation specialization trade-off} inherent to multi-task learning with a shared encoder. Four mechanisms explain the phenomenon:

\textbf{1. Severity is naturally encoded in pretrained representations.} Severity (L0--L4) corresponds to the overall intensity and emotional charge of text---a signal that is abundant in general-domain pretraining corpora (news, social media, literature). Pretrained language models learn to distinguish gradations of urgency, aggression, and distress without any task-specific supervision. A frozen encoder with a trainable linear classifier can therefore achieve $\sim$0.915 accuracy by reading these pre-existing intensity signals from the CLS embedding. This is analogous to how pretrained embeddings support strong linear-probe performance on sentiment analysis without fine-tuning.

\textbf{2. Fine-tuning reallocates representational capacity toward harm type discrimination.} \textsc{BRACE}'s training objective is dominated by fine-grained harm type classification: ORC provides four-step intermediate supervision over topic, indicator, severity, and type, in sequence, all oriented toward distinguishing among five harm categories. Prototype memory enriches features with category-level semantics. The shared encoder receives gradients that pull its representations toward harm type separability, which partially overwrites the severity-intensity signals that were already present in the pretrained embedding. H2 (gambling) already has the highest per-category F1 for every encoder linear probe in Table~\ref{tab:main} and remains the highest under BRACE; these results show an absolute improvement rather than a difficulty-ranking reversal.

\textbf{3. Gradient competition between harm type and severity objectives.} When the shared encoder updates its parameters to improve harm type classification, the resulting gradient direction is partially opposed to the direction that would improve severity classification, consistent with the multi-task trade-off observed in the available measured LP comparisons.

\textbf{4. Decoder models partially recover severity accuracy.} Qwen3-1.7B and ChineseGuard-1.5B, with LoRA fine-tuning, achieve severity accuracy of 0.850--0.852, surpassing all encoder backbones by $\sim$2.5 percentage points. This is consistent with the broader trend of decoder models outperforming encoders on tasks requiring nuanced semantic understanding: the larger parameter budget and richer pretrained representations allow decoders to maintain severity discrimination alongside harm type classification, though they still fall short of the linear probe ceiling (0.915--0.917).

\subsection{Implications}

This analysis reveals a fundamental tension in multi-task harmful content detection: \textbf{representations that optimally separate fine-grained harm categories are not the same representations that optimally capture severity intensity}. The linear probe upper bound (0.917) indicates that severity is inherently an ``easier'' task given pretrained features, but joint optimization forces a trade-off. Two architectural directions could address this: (1) a fully separate encoder tower for severity to eliminate gradient competition entirely, at the cost of doubled parameters; or (2) severity-specific feature adapters that learn a lightweight transformation from harm-type-oriented features back to severity-oriented features, preserving shared parameters while enabling task-specific readout. We leave exploration of these directions to future work and report severity as a secondary metric throughout the main paper, emphasizing that severity accuracy under \textsc{BRACE} should be interpreted as the cost of optimizing for fine-grained harm type discrimination rather than an absolute performance ceiling.

\section{Paired Context Sensitivity Diagnostic}
\label{sec:paired_context}

We evaluate 60 constructed dialogue pairs (120 dialogues), with 30 pairs each in Chinese and English, spanning five topics and three context patterns: reported quotation, expressed stance, and cross-turn reference. Within each pair, the final utterance is identical, while the preceding context distinguishes reporting or opposition from endorsement or recruitment. Binary labels are author-assigned and have not undergone independent review. This is an exploratory diagnostic of context sensitivity, not an estimate of performance on naturally occurring conversations.

Each saved model is evaluated with the full dialogue and with only the final utterance, using a fixed binary threshold of 0.5. Pairwise correctness requires both members of a pair to be classified correctly. Benign false-positive rate (FPR) is the fraction of benign members predicted harmful; harmful recall is the fraction of harmful members predicted harmful. Binary macro F1 averages the F1 scores for the two binary labels. Table~\ref{tab:paired_context} is recomputed from the archived per-dialogue probabilities. Initial and follow-up checkpoints are reported separately; the binary router supplies predictions in both cases, so this diagnostic does not test binary prediction fusion.

\begin{table*}[t]
\centering
\footnotesize
\setlength{\tabcolsep}{2mm}
\begin{tabular*}{\textwidth}{@{\extracolsep{\fill}}llcccc}
\toprule
\textbf{Variant} & \textbf{Input} & \textbf{Pair correct} & \textbf{Benign FPR} & \textbf{Harmful recall} & \textbf{Binary m-F1} \\
\midrule
$B$ & Full & 0.0333 & 0.5500 & 0.3833 & 0.4160 \\
$B$ & Last only & 0.0000 & 0.4833 & 0.4833 & 0.4999 \\
$P_1$ & Full & 0.0000 & 0.5500 & 0.4667 & 0.4583 \\
$P_1$ & Last only & 0.0000 & 0.5000 & 0.5000 & 0.5000 \\
$O_0$ (initial) & Full & 0.0000 & 0.7000 & 0.5333 & 0.4086 \\
$O_0$ (initial) & Last only & 0.0000 & 0.4500 & 0.4500 & 0.4987 \\
$O_1$ (initial) & Full & 0.0000 & 0.6667 & 0.5167 & 0.4201 \\
$O_1$ (initial) & Last only & 0.0000 & 0.5167 & 0.5167 & 0.4999 \\
$N$ & Full & 0.0000 & 0.7167 & 0.5833 & 0.4203 \\
$N$ & Last only & 0.0000 & 0.4833 & 0.4833 & 0.4999 \\
$O_0$ (follow-up) & Full & 0.0167 & 0.6167 & 0.5000 & 0.4398 \\
$O_0$ (follow-up) & Last only & 0.0000 & 0.4833 & 0.4833 & 0.4999 \\
$O_1$ (follow-up) & Full & 0.0167 & 0.6667 & 0.5333 & 0.4276 \\
$O_1$ (follow-up) & Last only & 0.0000 & 0.4833 & 0.4833 & 0.4999 \\
\bottomrule
\end{tabular*}
\caption{Exploratory context diagnostic on 60 constructed pairs. Higher pairwise correctness, recall, and macro F1 are better; lower benign FPR is better. Each row uses all 120 dialogues.}
\label{tab:paired_context}
\end{table*}

\noindent\textbf{Finding:} Full-context pairwise correctness ranges from 0 to 0.0333, with benign FPR from 0.5500 to 0.7167. The low pairwise correctness indicates frequent errors in distinguishing reporting or opposition from endorsement or recruitment in these constructed dialogues. With final-utterance-only input, the two members of each pair are identical and receive the same decision, so pairwise correctness is necessarily zero. This structural property alone is not evidence that adding context helps. The three context patterns share anchors, so pairs are not independent observations; we make no independent-pair significance claim.

\section{Blend-Ratio Sensitivity}
\label{sec:blend_sensitivity}

Table~\ref{tab:blend_sensitivity} reports archived ERNIE-3.0-Medium evaluations for five direct/chain logit weight ratios on the original 9,000-example test set. The available training summaries for 65/35, 70/30, 75/25, and 80/20 record five training epochs per run. A corresponding training summary for 60/40 is unavailable, so its training budget cannot be confirmed. These are individual archived runs, not three-seed averages, and their absolute scores should be distinguished from the main multi-seed comparison.

\begin{table}[t]
\centering
\footnotesize
\setlength{\tabcolsep}{1mm}
\begin{tabular*}{\columnwidth}{@{\extracolsep{\fill}}lccc}
\toprule
\textbf{Direct/chain} & \textbf{Epochs} & \textbf{Harm m-F1} & \textbf{Sev. Acc.} \\
\midrule
60/40 & --- & 0.9422 & 0.8167 \\
65/35 & 5 & 0.9415 & 0.8269 \\
70/30 & 5 & 0.9401 & 0.8254 \\
75/25 & 5 & 0.9399 & 0.8254 \\
80/20 & 5 & 0.9402 & 0.8229 \\
\bottomrule
\end{tabular*}
\caption{Archived blend-ratio evaluations. A dash denotes an unavailable training summary.}
\label{tab:blend_sensitivity}
\end{table}

\noindent\textbf{Finding:} Among runs with a confirmed five-epoch budget, harm macro F1 ranges from 0.9399 to 0.9415. The 70/30 configuration lies within this narrow observed range, but is not the highest-scoring ratio in this sweep. The 60/40 record has harm macro F1 of 0.9422 with lower severity accuracy (0.8167); its incomplete training record limits direct comparison. Without repeated runs, these differences do not establish a statistically preferred ratio.

\end{document}